\pdfoutput=1
\documentclass{article}

     \PassOptionsToPackage{numbers, compress}{natbib}

     \usepackage[preprint]{neurips_2019}

\usepackage[utf8]{inputenc} 
\usepackage[T1]{fontenc}    
\usepackage{hyperref}       
\usepackage{url}            
\usepackage{booktabs}       
\usepackage{amsmath}
\usepackage{amsfonts}       
\usepackage{nicefrac}       
\usepackage{microtype}      
\newtheorem{theorem}{Theorem}
\usepackage[pdftex]{graphicx}  
\usepackage{multirow}

\title{Graph Attention Memory for Visual Navigation}

\author{%
  Dong~Li, Qichao~Zhang, Dongbin~Zhao\thanks{Corresponding Author} \\
  The State Key Laboratory of Management and Control for Complex Systems, \\
  Institute of Automation, Chinese Academy of Sciences, Beijing, China\\
  University of Chinese Academy of Sciences, Beijing, China\\
  \texttt{\{lidong2014,zhangqichao2014,dongbin.zhao\}@ia.ac.cn} \\
  \And
  Yuzheng~Zhuang, Bin~Wang, Wulong~Liu, Rasul~Tutunov, Jun~Wang\\
  Huawei Noah's Ark Lab \\
  \texttt{\{zhuangyuzheng,wangbin158,liuwulong,rasul.tutunov,w.j\}@huawei.com} \\
}

\begin{document}

\maketitle

\begin{abstract}
Visual navigation in complex environments is inefficient with traditional reactive policy or general-purposed recurrent policy. To address the long-term memory issue, this paper proposes a graph attention memory (GAM) architecture consisting of memory construction module, graph attention module and control module. The memory construction module builds the topological graph based on supervised learning by taking the exploration prior. Then, guided attention features are extracted with the graph attention module. Finally, the deep reinforcement learning based control module makes decisions based on visual observations and guided attention features. Detailed convergence analysis of GAM is presented in this paper. We evaluate GAM-based navigation system in two complex 3D environments.  Experimental results show that the GAM-based navigation system significantly improves learning efficiency and outperforms all baselines in average success rate.
\end{abstract}

\section{Introduction}
Recently, visual navigation has received a lot of attention and has been applied to many fields, e.g. the house-robot navigation\citep{zhu2017target,Chen2019ABA} and large city navigation\citep{mirowski2018learning}. In order to efficiently navigate to the goal, there are many aspects an agent needs to concern. Firstly, due to the rich information of the visual observation, the agent needs to extract essential features to facilitate the learning process. Secondly, the agent should be capable of locating itself and utilizing the temporal relationship at different timescales to move towards the target position. Finally, the agent needs to well understand the environment's layout to generalize to new initial and goal positions. These factors make visual navigation a challenging problem.

With the success of deep reinforcement learning (DRL) methods, remarkable performance has been achieved in the visual navigation task. The corresponding approaches can be broadly classified into two categories: firstly, the reactive methods \cite{zhu2017target} in which a feed-forward policy directly takes the observation and predicts the action; secondly, the general-purposed temporal memory based approach which determines the action by using a recurrent policy, e.g., the long short-term memory (LSTM) network \cite{mirowski2016learning}. However, these DRL methods have limitation in dealing with the long-term memory task which is a key issue in visual navigation. Due to the inherent architecture, the reactive policy predicts the action passively, and cannot remember the temporal relationship of consecutive observations and the environment's layout. Although the LSTM network can handle temporal features of consecutive observations, it can hardly propagate temporal features hundreds of steps away based on visual observations. Therefore, the general-purposed LSTM policy which was demonstrated empirically in \cite{savva2017minos} is also insufficient to handle the visual navigation task.

One way to alleviate the long-term memory issue is to explicitly introduce an external memory to represent the environment and facilitate the learning. In the robotics community, there are two popular ways to represent the environment, i.e., the metric map and the topological map \cite{thrun2005probabilistic}. The metric map-based methods \cite{gupta2017cognitive,zhang2017neural,parisotto2017neural} maintain a metric-based belief of the world which is updated and utilized to extract global features along with the learning process. The metric map typically divides the environment into fine-grained grid-world with each cell contains a precise metric measurement. However, in the visual navigation, accurate metric measurements may not always be necessary. Moreover, learning a good metric map itself is non-trivial\cite{gupta2017cognitive,Blukis2018FollowingHN}.

In order to build the external memory while not cause too much map representation learning overhead as in the metric map, the topological map is introduced to represent the environment as in \cite{savinov2018semi}. We rely on the intuition that when human-beings walk into a room, they locate themselves by utilizing some key landmarks, which provide  visual clues to guide the following moves. Hence we let the agent first do explorations and collect visual observations as priors. The key landmarks can be identified from the priors by measuring pairwise similarities. Then a topological graph memory \cite{savinov2018semi} would be built by using these key observations. Given this topological graph memory, we employ the soft-attention mechanism \cite{velivckovic2017graph} of graph convolutional network to recurrently extract guided attention features by concerning its neighbor nodes in the graph. We call this proposed topological graph memory with recurrent attention mechanism as the graph attention memory (GAM) and give a detailed convergence analysis about it. Finally, we build a DRL navigation system by integrating the GAM into the agent. Note the proposed GAM module is independent to the DRL methods. As a result, it can be plugged into any DRL methods as an auxiliary module to facilitate visual navigation.

We evaluate our method for the goal-seeking visual navigation task in two complex 3D mazes of ViZDoom \cite{Kempka2016ViZDoom}. The agent is randomly spawned at several positions in the maze with a goal image provided. It needs to determine the optimal actions for reaching the goal based on its current visual observation and guided features extracted by GAM, hence motivating to learn the goal-directed knowledge of the environment layout from the topological memory. The experimental results show that the proposed method outperforms the baselines both in the learning speed and the learning performance. Additionally, the model can also generalize to novel initial or goal positions.

\section{Related Work}
There are plenty of literature for the visual navigation task. Here we focus on some of the related works. Recently, the methods which combine deep neural networks and reinforcement learning have achieved remarkable performance on many tasks such as ATARI games \cite{mnih2015human}, the game of Go \cite{silver2016mastering}, autonomous driving \cite{li2017torcs}, and vehicle classification \cite{zhao2017deep}. In \cite{zhu2017target}, a siamese CNN based reactive policy is proposed to control an agent to transfer among different goals. \cite{pathak2017curiosity,Shao2018VisualNW} also employ a reactive policy to navigate in a three-dimensional maze. In addition, some works employ the general-purposed LSTM networks to deal with the temporal relationship. For example, several LSTM modules are stacked in the policy network to enhance the temporal memory in \cite{mirowski2018learning,pathak2018zero}. Although these approaches can reach the goal, but the training process is relatively long due to the long-term memory issue in visual navigation.

\begin{figure}
	\centering
	\includegraphics[trim= 180 150 220 100, clip, width=12 cm]{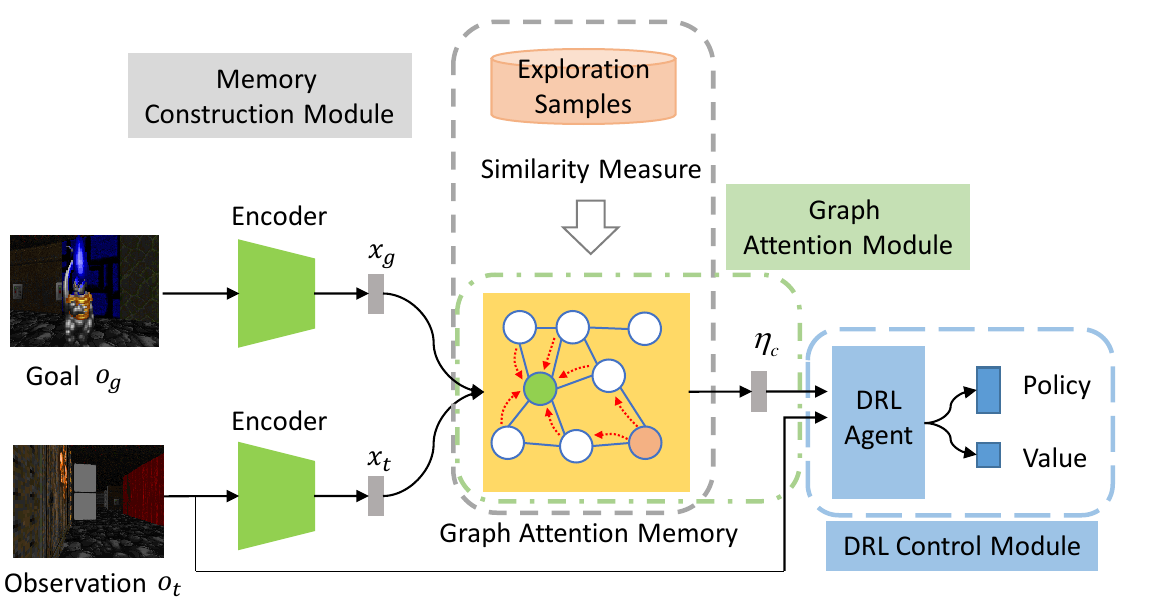}
	\caption{The visual navigation system overview. The system includes three modules: the memory construction module, the graph attention module, and the DRL control module.}
	\label{fig_system_framework}
\end{figure}

To alleviate the long-term memory issue, the external memory has been employed in many recent works to address the robot navigation task. The metric representation and the topological representation are two popular ways to describe the world in the robotics community. For the metric-based methods, a global egocentric top-view metric map is employed to predict free space and the corresponding confidence in \cite{gupta2017cognitive}. In \cite{parisotto2017neural,zhang2017neural}, a differential neural computer \cite{graves2016hybrid} is used to update the metric map and retrieve the hidden features. In \cite{chen2018learning}, the 3D points in the depth image are projected into a 2D grid world to form the metric map. \cite{Blukis2018FollowingHN} also uses the 3D to 2D projection approach to update the map, but they project the points from the final-layer feature map of a convolutional neural network (CNN) instead. An alternative way to represent the world is by using the topological map. Authors in \cite{gupta2017unifying} regard the demonstration samples as the key landmarks and construct a topological path description using the landmark images and the corresponding actions. The agent is trained based on the imitation learning. Different from their work, we build the topological map by using aimless exploration samples instead of the expert's task-orientated demonstration samples. Besides, the learning approach in this paper is based on the reinforcement learning rather than the imitation learning. The work of \cite{savinov2018semi} is more related to ours. They also build a topological map using aimless exploration samples. However, our method differs in the navigation framework and the way to use the topological map. The work in \cite{savinov2018semi} explicitly searches the shortest path by using Dijkstra's algorithm, then reach the intermediate waypoints one-by-one based on a supervised learning model. In this paper, we regard the topological map as an environment prior and extract guided features using the GAM module to facilitate reinforcement learning without any supervised signal for policy training. In addition, the GAM module is relatively independent to the DRL framework and can be used in other DRL approaches and the topological graph-based tasks.

\section{Methodology}
In this section, we detail the proposed visual navigation approach. The whole system includes three modules: the topological graph memory construction module, the guided attention feature extraction module, and the DRL control module as is shown in Figure \ref{fig_system_framework}. It works as following: 1) An exploration database is collected by manually or randomly control the agent to explore the environment. A binary classification neural network is employed to predict the sample similarities based on  the topological graph memory. 2) The agent is located based on the similarity probabilities between the current observation and the node features. The goal is also located similarly based on the visual similarity. By recurrently aggregating the neighbor node features, the guided attention features of the current node and goal node are obtained. 3) The DRL agent accepts the latent features from observation and the guided attention features to make the decision.

\subsection{Memory Construction Module}\label{sec_memory_construct}
This module attempts to construct the topological graph $ \mathcal{G}(\mathcal{V}, \mathcal{E}) $ of the environment based on the exploration samples. Since the exploration sample itself is a visual description of the corresponding environment region, the primitive method is to extract the visual features from the sample by using a CNN encoder. The node is represented by the visual features of the encoder output. Then a topological graph can be obtained by connecting the nodes of the consecutive exploration samples. However, the direct connection method has the limitation that the graph might be insufficient to describe the environment. For example, if two images are not observed consecutively but close in the environment, then they will not be connected. But they should be connected so as to give a visual description of the place where they are both located in.

With respect to the latent node connection issues above, an alternative approach is to employ a classification CNN to predict the connection probability of two observations \cite{savinov2018semi} besides the direct connection. In details, given two candidate observations $ o_i, o_j $, the classification CNN $ \phi(.,.;\theta_g) $ predicts the corresponding connection probabilities, i.e. $ p_{i,j} = \phi(o_i, o_j; \theta_g) $. Here $ \theta_g $ are the network weights. Since the graph edges only represent the connectivity rather than the connection weights, so the underlying network aims to deal with the binary classification task. While the learning task of the node connection problem is relatively simple, the key is to generate the valid training samples effectively. Assume the observation at time $ t $ is $ o_t $ in the exploration phase. In order to remove the trivial consecutive samples and consider relatively similar samples, the label $ y_{t,k} = 1 $ for the future sample $ o_k $ falling in the horizon $ T_{\text{min}} \le k-t \le T_{\text{max}} $, otherwise $ y_{t,k} = 0 $. Denote the connection CNN prediction as $ \hat{y}_{i,j} = \phi(o_i, o_j; \theta_g) $. The network weights $ \theta_g $ can be optimized by minimizing the following cross-entropy loss
\begin{equation}\label{eq_graph_loss}
\mathcal{L}(\theta_g) = \frac{1}{M} \sum_{i} y_{i,j}\log(\hat{y}_{i,j}) + (1-y_{i,j})\log(1-\hat{y}_{i,j}),
\end{equation}
where $ M $ is the number of mini-batch samples.

After training the connection CNN $ \phi(.,.;\theta_g) $, the topological graph memory can be constructed by taking the exploration prior. Assume the exploration sample database is $ \mathcal{D}_{exp}=\{o_i\}_{i=1}^{N} $ where $ N $ is the amount of exploration samples. The nodes in the graph can be determined by the exploration samples. Each node is represented by the embedding of each sample extracted after the last convolutional layer of network $ \phi(\cdot, \cdot; \theta_g) $. For a specific sample $ o_i \in \mathcal{D}_{exp} $, the probabilities of being connected with other nodes is computed by
\begin{equation}\label{eq_edge_probability}
p_i = \phi(o_i, o_k; \theta_g), \quad \forall k=1, ..., N.
\end{equation}
Therefore, the probability vector $ p_i \in \mathbb{R}^N$ where $ p_{i,j} $ is the connection probability of node $ i $ and $ j $. Given the threshold $ \epsilon $, the edge $ e_{i,j} $ of node $ i $ and $ j $ can be obtained by following
\begin{equation}\label{eq_connect_property}
e_{i,j} =
\begin{cases}
1, & \text{if $ p_{i,j} \ge \epsilon $,}	\\
0, & \text{otherwise}.
\end{cases}
\end{equation}
Based on the edges of all samples in $ D_{exp} $, the topological graph memory is then built. Note the graph is built only based on the observations without exploration actions. In this way, the graph is agent-agnostic which means the underlying physics dynamics of the agent is blocked away.

\subsection{Graph Attention Module}
This module aims to take advantage of the constructed graph memory above and extract essential guided features that can help the agent to reach the goal. To this end, the first step is to localize the agent and the goal on the graph. Since the node connection CNN $ \phi(o_i, o_j; \theta_g) $ in section \ref{sec_memory_construct} predicts the sample similarity, we can localize the agent in the graph by feeding the current observation into it and pick the most similar node as the current position. Note that in the localization process, the agent only uses the observation and does not have any access to the location ground-truth. The same operation can be used to localize the goal node. In practice, due to the environment texture similarity, selecting the node of maximum similarity will cause the localization jittering. To increase the localization robustness, we pick up top $ L $ similar nodes and select the node of median similarity score as the current node. After localization, the guided attention features are extracted as follows.

Assume there are $ N $ nodes in the graph node set $ \{x_i\}_{i=1}^N $ where $ x_i \in \mathbb{R}^D $ is the node feature vector extracted from the final convolutional layer of graph construction CNN in section \ref{sec_memory_construct}. For a node $ i $ and the node $ j $ of its neighbor $ \mathcal{N}_i $, the siamese attention network $ \psi(x_i, x_j; \theta_a) $ parameterized by weights $ \theta_a $ is employed to transfer them into the embedding space. According to \cite{wang2018non}, feeding the output of $ \psi(x_i, x_j; \theta_a) $ as the logit to the softmax function will obtain the pair-wise 
attention score in the embedding space. Therefore, the attention coefficient can be computed by following
\begin{equation}\label{eq_attention_score}
\alpha_{i,j} = \frac{e^{\psi(x_i, x_j; \theta_a)}}{\sum_{\forall k\in \mathcal{N}_i} e^{\psi(x_i, x_k; \theta_a)}}.
\end{equation}
The aggregated feature of node $ i $, i.e. $ x_i' $ can be obtained by taking an inner-product operation over the attention coefficients of $ \mathcal{N}_i $ and the corresponding node features
\begin{equation}\label{eq_aggregated_feature}
x_i' = \sum_{\forall j\in \mathcal{N}_i} \alpha_{i,j} x_j.
\end{equation}
Note that the neighbor $ \mathcal{N}_i = \{i\} \cup \{j: (i, j)\in \mathcal{E} \} $ includes the node $ i $ itself in order to keep its own feature in the aggregated feature $ x_i' $. Since we want the agent automatically learn where to focus on the graph, the attention network $ \psi(\cdot,\cdot;\theta_a) $ is trained end-to-end by maximizing the cumulative reward. The details are described in section \ref{sec_drl_control}.

\begin{figure}
	\centering
	\includegraphics[scale=0.45]{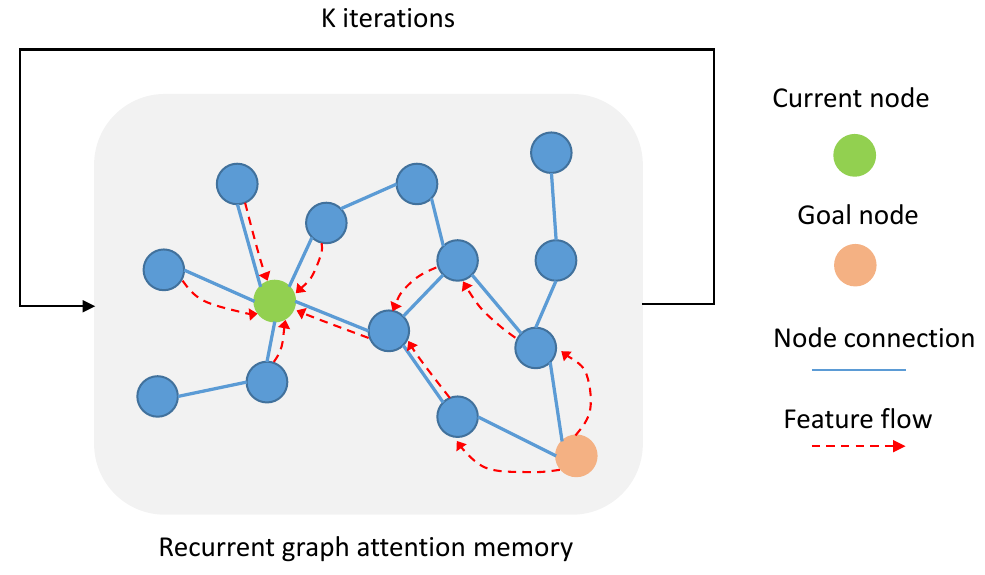}
	\caption{The graph attention memory with $ K $ recurrent iterations. After multiple recurrent operations, the goal node feature will spread and be aggregated into the current node through the feature flow.}
	\label{fig_graph_attention}
\end{figure}

Actually, the operation in Eq. \eqref{eq_aggregated_feature} is a one-step feature aggregation operation. A recurrent feature aggregation operation can be employed to obtain more stable aggregated features. We provide two ways to understand this. Firstly, we show this intuitively. Then, the convergence proof of the recurrent feature aggregation operation is formally given.

Typically the goal is hundreds of steps away from the initial node. The One-step feature aggregation in Eq. \eqref{eq_aggregated_feature} is insufficient to spread the temporal features between the current node and the goal node. To address this issue, a multi-step recurrent iteration mechanism over the graph memory is proposed as shown in Figure \ref{fig_graph_attention}. The red dashed lines represent several feature flow paths. Assume the aggregated node feature at  $ k $th iteration is $ x_{i}^{[k]} $, then the feature at  $ k+1 $th iteration is obtained by
\begin{equation}\label{eq_k_iter}
x_i^{[k+1]} = \sum_{\forall j\in \mathcal{N}_i} \alpha_{i,j} x_j^{[k]}.
\end{equation}
Repeating the above graph feature aggregation operation for multiple iterations, the latent features of the current node and the goal node will transfer to all other nodes in the graph. Note in Eq. \eqref{eq_k_iter}, only the original network weights $ \theta_a $ are used\footnote{The attention coefficient $ \alpha_{i,j} $ is a function of $ \theta_a $ as in Eq. \eqref{eq_attention_score}}, and no additional network weights are introduced.

Now we provide the \textit{recurrent graph aggregation theorem} to show that the recurrent feature aggregation operation in Eq. \eqref{eq_k_iter} will finally lead the node feature converge to its stationary value.

\begin{theorem}
	(Recurrent Graph Aggregation Theorem). Suppose that the stochastic matrix $ \boldsymbol{W} $ such that $ \boldsymbol{[W]}_{i,j} = \alpha_{i,j} $, $ i=1, \cdots N $, $ j = 1, \cdots N $, $ \alpha_{i,j} $ is computed according Eq. \eqref{eq_attention_score}. The matrix $ \boldsymbol{X} \in \mathbb{R}^{N\times D} $ is the node feature matrix in which each row $ [X]_i = x_i $ is the feature of node $ i $. Let $ \boldsymbol{X}^{[k]} $ be the $ k $th iteration of $ \boldsymbol{X} $. Then the limit of the recurrent graph aggregation matrix $ \boldsymbol{X}^* $ exists
	\begin{equation}\label{eq_theorem_recurrent}
	\lim\limits_{k \rightarrow \infty} \boldsymbol{W} \boldsymbol{X}^{[k]} = \boldsymbol{X}^{*}
	\end{equation}
\end{theorem}
The detailed proof is given in appendix\ref{appendix_theorem_proof}. It follows that the recurrent graph aggregation in Eq. \eqref{eq_k_iter} can be formulated as a random walk on graph $ \mathcal{G} $ where the stochastic matrix $ \boldsymbol{W} $ is the state transition probability matrix. Since $ \boldsymbol{W} $ is irreducible and aperiodic, then the stationary state matrix $ \boldsymbol{X}^{*}  $ of Markov chain exists and the chain converges to it.

Denote the aggregated features for the current node and goal node after the $ K $-th iteration as $ x_t^{(K)} $ and $ x_g^{(K)} $, respectively. We use the node feature difference as the guided attention features
\begin{equation}\label{eq_guided_attention_feature}
\eta_t = x_t^{(K)} - x_g^{(K)}.
\end{equation}
A key intuition that we rely on is that sensing the goal far away is more significant than nearby. Based on this, the guided attention feature $ \eta_t $ is bigger if the agent is far away from the goal, and vice versa. Combining Eq. \eqref{eq_attention_score}, \eqref{eq_k_iter} and \eqref{eq_guided_attention_feature}, the guided attention feature can be represented as $ \eta_t = \psi(x_t, x_g; \theta_a)$.

\subsection{DRL Control Module} \label{sec_drl_control}
After extracting the guided attention feature, We employ the synchronous advantage actor-critic (A2C) approach to optimize the agent network weights.

\textbf{Model.}
 The actor network $ \pi(s_t; \theta_\pi) $ accepts the state $ s_t $ and chooses the optimal action $ a_t $. The critic network $ V(s_t; \theta_v) $ also accepts the current state $ s_t $ and predicts the expected reward-to-go from $ s_t $ to the terminal state $ s_g $. Here $ \theta_v $ are the critic network weights. Define the discounted cumulative reward from time $ t $ is $ R_t = \sum_{k=t}^{T} \gamma^{k-t} r_k $ where $ \gamma \in (0, 1) $ is the discount rate and $ T $ is the terminal time step. According to the policy gradient theorem\citep{mnih2016asynchronous}, the actor network maximizes the expected discounted cumulative reward $ \log \pi(s_t; \theta_\pi) (R_t - V(s_t; \theta_v)) + \beta H(\pi(s_t; \theta_\pi))$, where $ H(.) $ is the entropy and $ \beta $ adjusts its strength. Here the entropy item is added to prevent the premature. The critic network minimizes the square error between $ R_t $ and $ V(s_t; \theta_v) $ to predict the state value accurately. Note the graph attention memory mechanism is a feature extraction procedure for the agent, i.e. $ s_t = \psi(o_t, o_g; \theta_a) $, so the network weights $ \theta_a $ are jointly optimized with $ \theta_\pi $ and $ \theta_v $. Above all, the agent optimizes all the weights by minimizing the loss
\begin{align}\label{eq_rl_loss}
\mathcal{L}(\theta_\pi, \theta_v, \theta_a) & = \frac{1}{T_h} \sum_{i=1}^{T_h} -\log \pi(s_t; \theta_\pi) (R_t - V(s_t; \theta_v)) \nonumber \\
& + (R_t - V(s_t; \theta_v))^2 - \beta H(\pi(s_t; \theta_\pi)).
\end{align}

\begin{figure}
	\centering
	\includegraphics[trim=130 200 110 170, clip, width=14 cm]{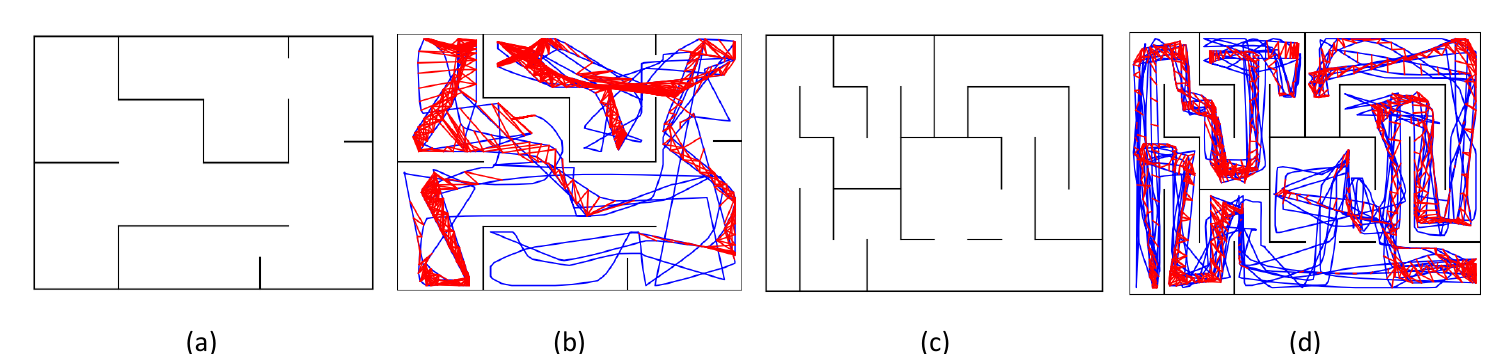}
	\caption{Two maze layouts (a), (c) and the corresponding exploration trajectories (blue lines) and node edges (red lines) in (b), (d).}
	\label{fig_mazelayout}
\end{figure}

\section{Experiments}
In this section, we validate the GAM-based navigation in two aspects, which include the quality of the constructed memory and the effectiveness of the guided attention features. We give the experimental setup firstly and detail the experiment results later.

\subsection{Experiment Setup}
We implement all experiments on a computer with an Intel Xeon E5-2620 CPU and Nvidia Titan Xp GPU. To build the graph memory, we employ the shared ResNet-18 \cite{he2016deep} as the encoder to extract features for two observations. The features are then concatenated and fed into five fully connected (FC) layers . Each FC layer includes 512 hidden units excepting the fifth layer has 2 outputs for the binary classification. For the graph attention module, the policy and value share three convolutional layers (Conv(ch=32, k=8, s=4), Conv(ch=64, k=4, s=2), Conv(ch=32, k=3, s=1)) and one fully connected (FC) layer with 256 hidden units where ch is the number of kernels, k is the kernel size, and s is the stride. After the last FC layer, a separated FC layer outputs the policy $ \pi(s, a) $ and the value $ V(s) $. For the attention module, it includes three convolutional layers (Conv(ch=16, k=1, s=1), Conv(ch=1, k=1, s=1), Conv(ch=1, k=1, s=1)) and a FC layer with 16 hidden units. To increase the network robustness, we use multiple attention modules and the outputs are concatenated to form the guided attention vector $ \eta_t $. The guided attention vector and observation features are concatenated at the first FC layer of the policy network. The horizon $ T_{\text{min}}=5, T_{\text{max}} =20.$ The node connection threshold $ \epsilon $ is determined by the top-$ L $ probability where $ L $ is the edges computed by Eq. \eqref{eq_connect_property}.

The state is the combination of the current observation and the graph attention features, i.e. $ s_t = [o_t, \eta_t] $. The action space includes seven discrete actions \{\text{move\_forward}, \text{move\_backward}, \text{move\_left}, \text{move\_right}, \text{turn\_left}, \text{turn\_right}, \text{not\_move}\}. The $ \epsilon $-greedy strategy is used to select the action. The agent will obtain the reward of 10.0 if it reaches the goal. Otherwise, a small penalty of -0.05 is given in every step.

The training protocol is as follows. For the memory construction network, we use the adam optimizer with learning rate $ 0.001 $. The mini-batch size is 64, and the network is trained for 10000 epochs. For the GAM-based navigation, we use the RMSProp optimizer with learning rate $ 2.5e^{-4} $. The network input is gray-scale image with size $ 84 \times 84 $. An episode only terminates when the time step arrives 2000. If the agent reaches the goal, it will be randomly re-spawned in the maze. We totally train the agent for $ 5e^7 $ steps.

\subsection{Memory Construction Results}
To train the memory construction network, we build two mazes with different complexities in  ViZDoom as shown in Figure \ref{fig_mazelayout}. The  size of maze1 in Figure \ref{fig_mazelayout} (a) is $ 768 \times 768 $ while the size of maze 2 in Figure \ref{fig_mazelayout} (c) is $ 1280 \times 640 $.  In order to deal with over-fitting and increase the data diversity, we change the maze wall textures while keep the maze layout invariant. Finally, $ 400 $ different mazes are built in total for each layout. In order to validate the effectiveness of the built memory, we manually control the agent to aimlessly explore the mazes. After computing the connection probabilities of all nodes, we select $ L=1300 $ and $ L = 1800 $ connections with highest probabilities for maze1 and maze2, respectively. The corresponding built graph memories are shown in Figure \ref{fig_mazelayout} (b) and (d). The blue lines represent the agent exploration trajectories and the red lines are the node connection. It can be seen that most nodes are connected with their nearby neighbors. Additionally, there are no wrong connections like the too distant connection and the connections which through the wall. 

\begin{figure}
	\centering
	\includegraphics[trim=120 120 120 140,clip,width=13 cm]{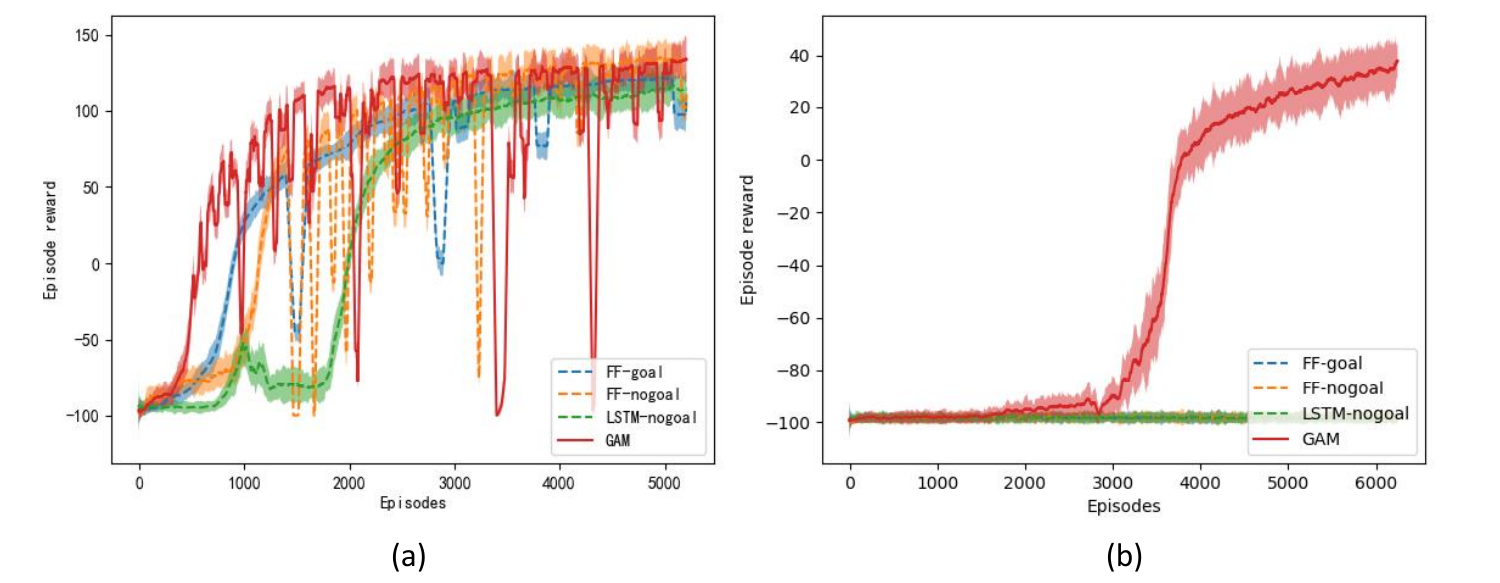}
	\caption{The training rewards of an episode in (a) maze1 and (b) maze 2. The solid lines are GAM-based method and the dashed lines are baselines.}
	\label{fig_all_reward}
\end{figure}

\begin{figure}
	\centering
	\includegraphics[trim=120 180 120 140,clip,width=12 cm]{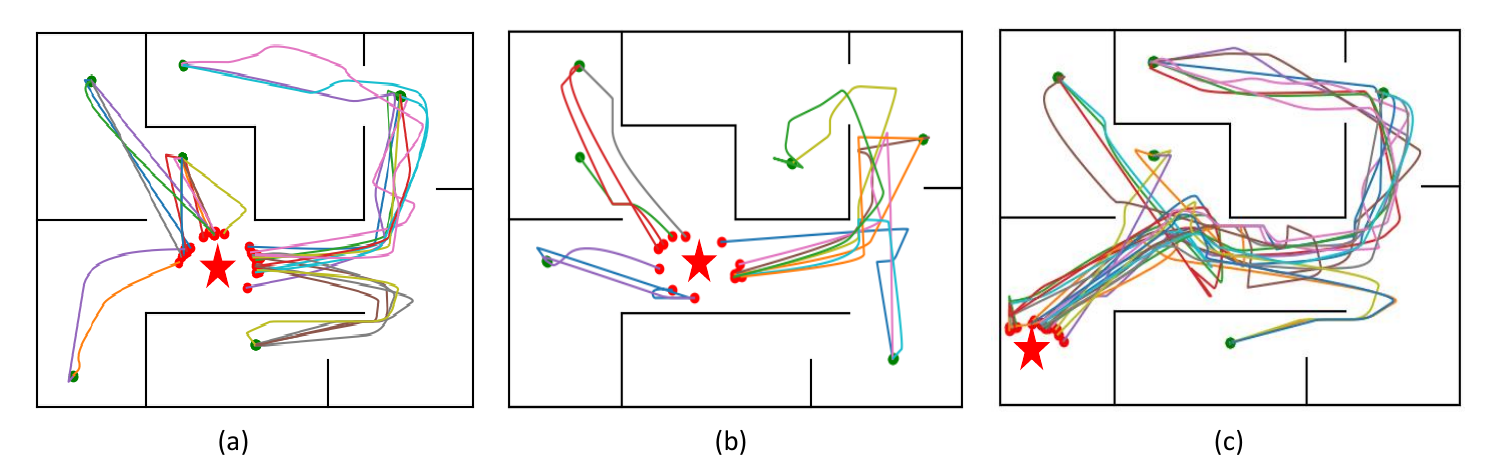}
	\caption{The trajectories of the GAM agent in an episode on maze 1. The green and red circles represent the initial positions and the goal position. (a) shows the trajectories with the training initial positions and goal positions; (b) and (c) show the performance of the trained model to the novel initial and goal positions.}
	\label{fig_traj}
\end{figure}

\subsection{GAM-based Navigation Results}
Since the GAM module is intended to facilitate the learning process of the DRL agent, we compare it with several baselines to validate the performance of the proposed GAM-based navigation method. \textit{(1) FF-nogoal}: the feed-forward architecture with three convolutional layers; \textit{(2) FF-goal}: the feed-forward architecture but also includes the goal image in the state; \textit{(3) LSTM-nogoal}: the same as the FF-nogoal baseline except the last fully connected layer is replaced by the LSTM layer with 256 hidden units. The policy and value network architectures of all baselines are the same as the GAM agent's. The training rewards of an episode are shown in Figure \ref{fig_all_reward}. It can be seen that the GAM agent can not only learns faster but converge to a higher episode reward than all the baselines. The GAM agent performs significantly better on maze 2 which is much more complex and difficult than maze 1. The reason is that the goal is much far away from the initial positions in maze 2 causing the baselines hard to reach the goal by chance and obtain no positive feedbacks. Since the episode reward is the total received reward in a fixed time interval, i.e. 2000, so the GAM agent can reach the goal more often. Thus, the GAM agent can reach the goal more efficient than the baselines. 

The testing scores and success rate of the GAM agent and the baselines are reported in Table \ref{tb_testing_score}. The total steps are 2000 for maze 1 and 6000 for maze 2. The success rate is measured by initializing the agent at all initial positions and checking whether it can reach the goal with 500 steps. Both the baselines and the GAM agent can successfully reach to the goal on maze 1. But GAM agent can achieve 100\% navigation success rate while the baselines can only reach the goal from the closest initial position on maze 2. This further validates the effective of the graph attention feature.

\begin{table}
	\caption{The testing scores and success rate of the GAM-based navigation and the baselines.}
	\label{tb_testing_score}
	\centering
	\begin{tabular}{llllll}
		\toprule
                                            &                            & FF-nogoal & \multicolumn{1}{l}{FF-goal} & \multicolumn{1}{l}{LSTM-nogoal} & \multicolumn{1}{l}{GAM} \\
                                            \hline
\multicolumn{1}{c}{\multirow{2}{*}{Maze 1}} & \multicolumn{1}{c}{Scores} & 122.05    & 122.05                      & 126.05                          & \textbf{132.05}                         \\
\multicolumn{1}{c}{}                        & Success Rate               & 100\%     & 100\%                       & 100\%                           & \textbf{100\%   }                       \\
\hline
\multirow{2}{*}{Maze 2}                     & \multicolumn{1}{c}{Scores} & -289.95   & -299.95                     & -299.95                         & \textbf{100.5}                          \\
& Success Rate               & 10\%      & 10\%                        & 10\%                            & \textbf{100\%}          \\
		\bottomrule
	\end{tabular}
\end{table}

We also show the trajectories of the GAM agent in an episode in Figure \ref{fig_traj}. The green and red circles represent the initial positions and the goal position, respectively. It shows that the agent can reach the goal smoothly without jittering actions. Additionally, the generalization performance of the trained model is reported. Figure \ref{fig_traj} (b) shows the generalization to the new initial positions. All six initial positions are changed while the goal position keeps the same as the training settings. The trained gam-nogoal agent can successfully navigated to the goal. Figure \ref{fig_traj} (c) represents the goal position changing case in which the agent can also robustly generalize to the new goal position. The navigation trajectories on the maze 2 are reported in the supplement material.

\section{Conclusion}
In this paper, we propose a graph attention memory (GAM) based approach for the visual navigation task. To address the long-term memory limitation of the reactive deep reinforcement learning  policy, GAM module aims to recurrently extract guided attention features based on the exploration prior. The whole GAM-based navigation system includes three modules: the topological graph memory construction, the guided attention feature extraction, and the DRL control. 

We theoretically analyze the convergence property of the proposed recurrent guided attention extracting operation. The experiments show that the GAM agent can achieve faster learning speed and higher success rate compared to the baselines. We also visualize the built graph memory to give a qualitative validation. The results show that there are no wrong connections between distant nodes and nodes separated by the wall. The navigation trajectories show that the agent can reach the goal smoothly without jittering actions. Additionally, the trained agent can generalize to the unknown initial and goal positions.

Our future work includes employing the dynamic graph construction mechanism rather than the static topological graph. Additionally, we aim to transfer the GAM-based navigation system from the simulator to the real-world robot visual navigation task.

\newpage
{
\small
\bibliographystyle{plainnat}
\bibliography{mybib}
}
\newpage
\appendix
\normalsize

\section{Convergence Proof of Theorem1}\label{appendix_theorem_proof}
Suppose that the stochastic matrix $ \boldsymbol{W} $ such that $ \boldsymbol{[W]}_{i,j} = \alpha_{i,j} $, $ i=1, \cdots N $, $ j = 1, \cdots N $, $ \alpha_{i,j} $ is computed according to 
\begin{equation}
\alpha_{i,j} =
\begin{cases}
\frac{\exp{\psi(\boldsymbol{x}_i,\boldsymbol{x}_j, \boldsymbol{\theta}_a)}}{\sum_{k\in\mathcal{N}_i}\exp{\psi(\boldsymbol{x}_i,\boldsymbol{x}_k, \boldsymbol{\theta}_a)}} & \text{if $j\in\mathcal{N}_i$} \\
0 & \text{otherwise} 
\end{cases}
\end{equation}
where $\mathcal{N}_i = \{i\}\cup \{j: (i,j) \in\mathcal{E}\}$ is a neighbourhood of node $i$ in $\mathcal{G}$ (including node $i$). Please notice, that $\boldsymbol{W}$ is a stochastic matrix, corresponding to random walk on graph $\mathcal{G}$.

Assume that we start with some feature vector $\boldsymbol{x}^{[0]}_i$ for each observation $\boldsymbol{o}_t$. Let us consider the following update mechanism:
\begin{equation}\label{eq_1}
\boldsymbol{x}^{[k+1]}_i = \sum_{j\in\mathcal{N}_i}\alpha_{i,j}\boldsymbol{x}^{[k]}_j \ \ \ i = 1,\ldots, N
\end{equation}

Let us denote the $\ell^{th}$ component of vector$\boldsymbol{x}^{[k]}_i$ as $\boldsymbol{x}^{[k]}_i(\ell)$, then, the above updating rule can be decomposed to the collection of univariate  updates: 
\begin{equation}\label{eq_2}
\boldsymbol{x}^{[k+1]}_i(\ell) = \sum_{j\in\mathcal{N}_i}\alpha_{i,j}\boldsymbol{x}^{[k]}_j(\ell) \ \ \ i = 1,\ldots, N \ \ \ \text{ and } \ell = 1,\ldots D. 
\end{equation}
where $D$ is the dimmensionality of the feature vector $\boldsymbol{x}_t$. To study the convergence properties of equation (\ref{eq_2}) (and respectively (\ref{eq_1})) let us introduce for each $\ell =1,\ldots,D$, a variable $\boldsymbol{y}^{[k]}_{\ell} = \left[\boldsymbol{x}^{[k]}_1(\ell),\boldsymbol{x}^{[k]}_2(\ell),\ldots, \boldsymbol{x}^{[k]}_N(\ell) \right]^{\mathsf{T}}$. In other words, in vector $\boldsymbol{y}^{[k]}_{\ell}$ we aggregate $\ell^{th}$ components of all vectors $\boldsymbol{x}^{[k]}_1, \ldots, \boldsymbol{x}^{[k]}_N$. It is easy to see, that equation (\ref{eq_2}) then can be written as:
\begin{align}\label{eq_3}
&\boldsymbol{y}^{[k+1]}_1 = \boldsymbol{W}\boldsymbol{y}^{[k]}_1\\\nonumber
&\vdots\\\nonumber
&\boldsymbol{y}^{[k+1]}_{\ell}= \boldsymbol{W}\boldsymbol{y}^{[k]}_{\ell}\\\nonumber
&\vdots\\\nonumber
&\boldsymbol{y}^{[k+1]}_D = \boldsymbol{W}\boldsymbol{y}^{[k]}_D\\\nonumber
\end{align}
which can be considered as independent sequence of $d$ averaging protocols, implemented on the same set of states $\mathcal{V}$. Each of this averaging protocol has its own initial state  $\boldsymbol{y}^{[0]}_{\ell}$\footnote{Each starting distribution follows immediately from $\boldsymbol{x}^{[0]}_i$ with $i=1,\ldots, N$.} but all of them defined the same asymmetric transition matrix $\boldsymbol{W}$. Moreover, due to the construction of $\boldsymbol{W}$ the corresponding underlying dynamics is irreducible and aperiodic. Hence, the spectral radius $\boldsymbol{W}$ is $1$ and there is exactly one eigenvalue on unit disk. Let $\boldsymbol{\pi}$ be the left Perron vector of $\boldsymbol{W}$ corresponding to the eigenvalue $1$, i.e $\boldsymbol{\pi}^{\mathsf{T}}\boldsymbol{W} = \boldsymbol{\pi}^{\mathsf{T}}$, then we have:
\begin{equation}\label{s_eq_w_lim}
\lim_{k\to\infty}\boldsymbol{W}^{k} = \boldsymbol{1\pi}^{\mathsf{T}}
\end{equation}
Hence, for all $\ell = 1,\ldots,D$ we have:
\begin{equation}\label{s_eq_y_lim}
\lim_{k\to\infty}\boldsymbol{y}^{[k]}_{\ell} = \boldsymbol{1\pi}^{\mathsf{T}}\boldsymbol{y}^{[0]}_{\ell}.
\end{equation}
Therefore, the corresponding $\boldsymbol{x}^{[k]}_{i}$ will also converge:
\begin{equation}
\lim_{k\to\infty}\boldsymbol{x}^{[k]}_i = \lim_{k\to\infty}\begin{bmatrix}
\boldsymbol{y}^{[k]}_1(i) \\
\boldsymbol{y}^{[k]}_2(i) \\
\vdots\\
\boldsymbol{y}^{[k]}_D(i)
\end{bmatrix} =  \begin{bmatrix}
\boldsymbol{\pi}^{\mathsf{T}}\boldsymbol{y}^{[0]}_1 \\
\boldsymbol{\pi}^{\mathsf{T}}\boldsymbol{y}^{[0]}_2\\
\vdots\\
\boldsymbol{\pi}^{\mathsf{T}}\boldsymbol{y}^{[0]}_D  
\end{bmatrix} .
\end{equation}
Define  $ \boldsymbol{X}^{[k]} = [\boldsymbol{y}_{1}^{[k]}, \cdots, \boldsymbol{y}_{D}^{[k]} ] \in \mathbb{R}^{N\times D} $, $ \boldsymbol{X}^* = [\boldsymbol{1\pi}^{\mathsf{T}}\boldsymbol{y}^{[0]}_{1}, \cdots, \boldsymbol{1\pi}^{\mathsf{T}}\boldsymbol{y}^{[0]}_{D}]  $. By according to Eq. \eqref{s_eq_w_lim} and \eqref{s_eq_y_lim}, then the theorem holds
\begin{equation}
\lim\limits_{k \rightarrow \infty} \boldsymbol{W} \boldsymbol{X}^{[k]} = \boldsymbol{X}^{*}.
\end{equation}

\section{GAM Navigation Trajectories on Maze 2}
\begin{figure}[!htbp]
	\centering
	\includegraphics[trim=120 120 120 140,clip,width=13 cm]{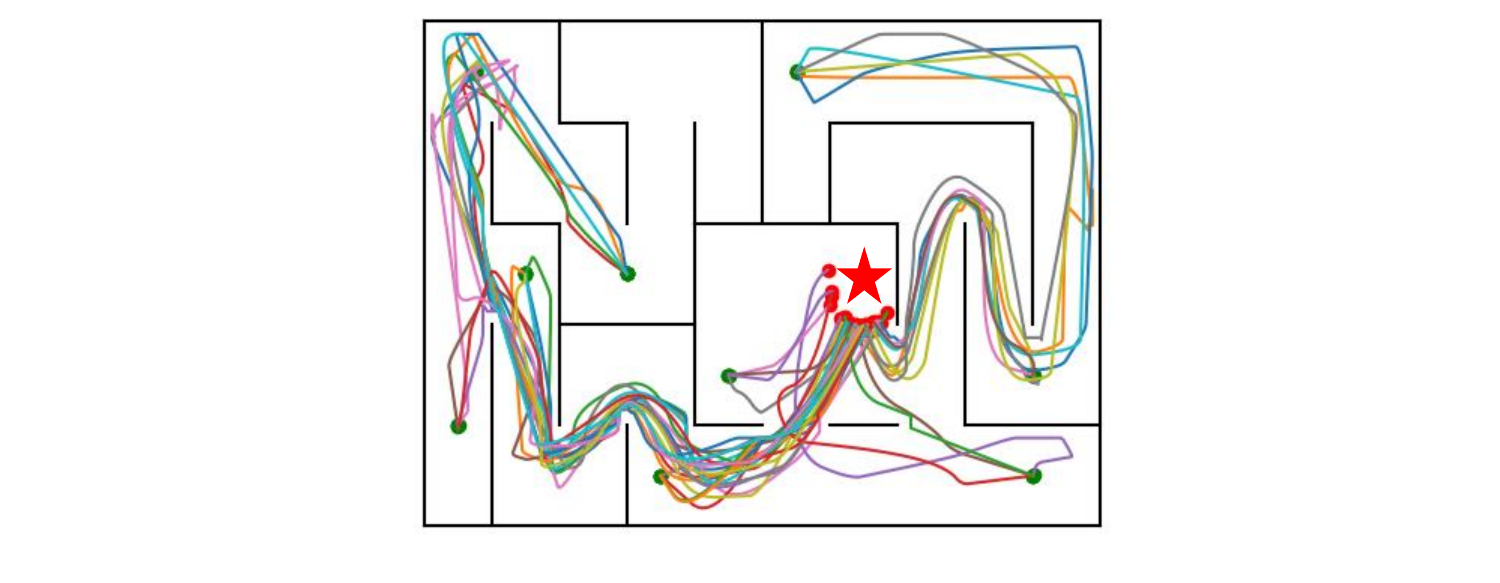}
	\caption{The GAM agent nagivation trajectories on maze 2.}
	\label{s_fig_trajectories_maze2}
\end{figure}

The GAM agent nagivation trajectories in 6000 steps on maze 2 are shown in Figure \ref{s_fig_trajectories_maze2}. The agent can efficient reach the goal without hover actions.

\begin{figure}[!htbp]
	\centering
	\includegraphics[trim=200 0 200 0,clip,width=14 cm]{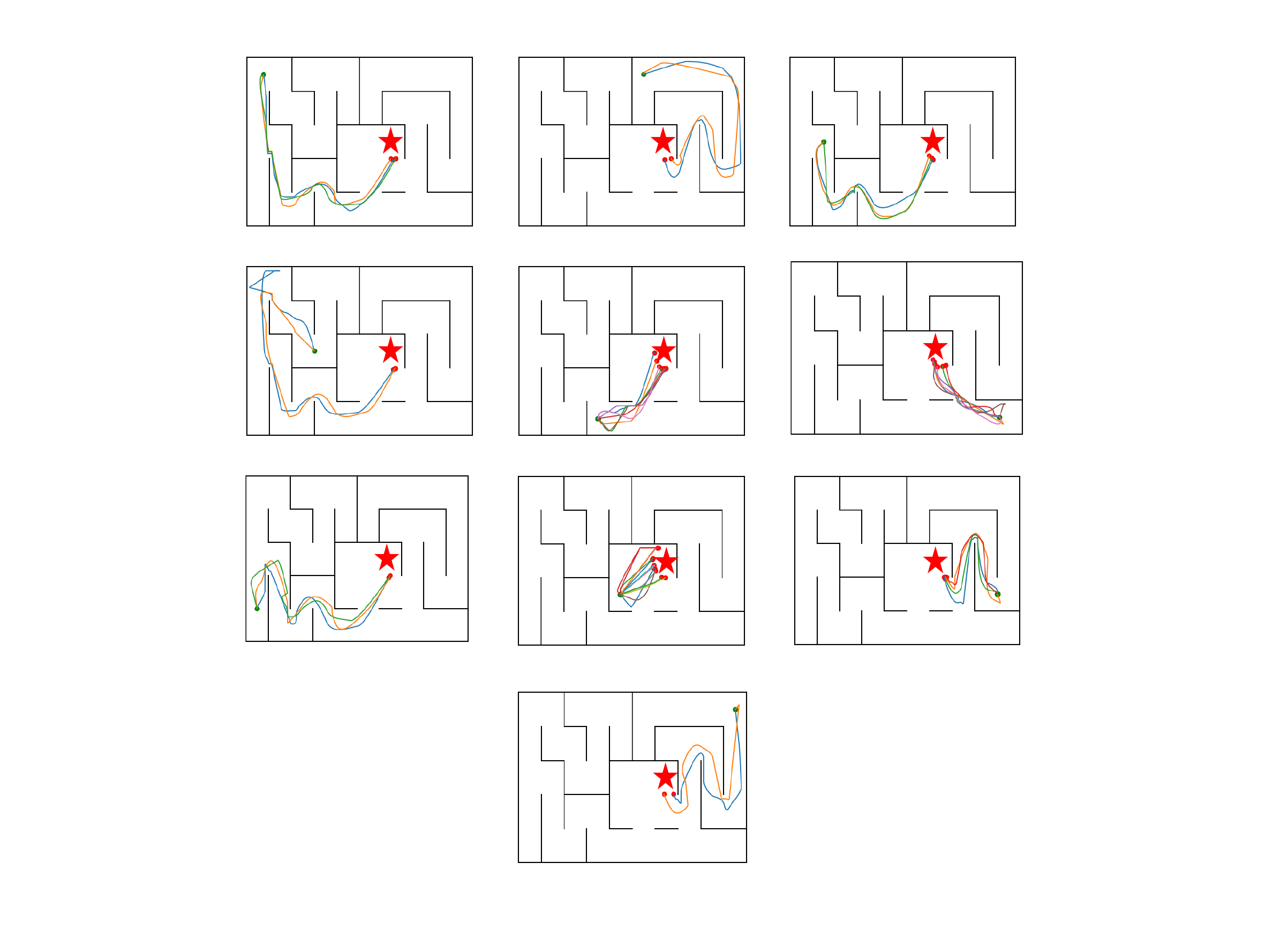}
	\caption{The GAM agent nagivation trajectories from ten initial  positions on maze 2.}
	\label{s_fig_traj_initial_position}
\end{figure}

Figure \ref{s_fig_traj_initial_position} shows the trajectories from all 10 initial positions to the goal on maze 2. It can be seen from the Figure \ref{s_fig_traj_initial_position} that the agent can smoothly navigate to the goal.

\end{document}